\documentclass{ieeeaccess}
\usepackage{cite}
\usepackage{amsmath,amssymb,amsfonts}
\usepackage{algorithmic}
\usepackage{graphicx}
\usepackage{textcomp}
\usepackage[utf8]{inputenc}
\usepackage{float}

\usepackage{bm}
\makeatletter
\AtBeginDocument{\DeclareMathVersion{bold}
\SetSymbolFont{operators}{bold}{T1}{times}{b}{n}
\SetSymbolFont{NewLetters}{bold}{T1}{times}{b}{it}
\SetMathAlphabet{\mathrm}{bold}{T1}{times}{b}{n}
\SetMathAlphabet{\mathit}{bold}{T1}{times}{b}{it}
\SetMathAlphabet{\mathbf}{bold}{T1}{times}{b}{n}
\SetMathAlphabet{\mathtt}{bold}{OT1}{pcr}{b}{n}
\SetSymbolFont{symbols}{bold}{OMS}{cmsy}{b}{n}
\renewcommand\boldmath{\@nomath\boldmath\mathversion{bold}}}
\makeatother

\def\BibTeX{{\rm B\kern-.05em{\sc i\kern-.025em b}\kern-.08em
    T\kern-.1667em\lower.7ex\hbox{E}\kern-.125emX}}

\begin{document}
\history{Date of publication xxxx 00, 0000, date of current version xxxx 00, 0000.}
\doi{10.1109/ACCESS.2024.0429000}

\title{Generating Explanations for Autonomous Robots: a Systematic Review}
\author{\uppercase{David Sobrín-Hidalgo}\authorrefmark{1},
\uppercase{Ángel Manuel Guerrero-Higueras}\authorrefmark{1}, and \uppercase{Vicente Matellán-Olivera\authorrefmark{1}}}

\address[1]{Robotics Group of University of León. Campus de Vegazana, 24071 León, Spain}

\tfootnote{This research has been partially funded by the Recovery, Transformation, and Resilience Plan, financed by the European Union (Next Generation) thanks to the TESCAC project (Traceability and Explainability in Autonomous Systems for improved Cybersecurity) granted by INCIBE to the University of León, and by grant PID2021-126592OB-C21 funded by MCIN/AEI/10.13039/501100011033 EDMAR (Explainable Decision Making in Autonomous Robots) project, PID2021-126592OB-C21 funded by MCIN/AEI/10.13039/501100011033 and by ERDF ''A way of making Europe''.}

\markboth
{Author \headeretal: Preparation of Papers for IEEE TRANSACTIONS and JOURNALS}
{Author \headeretal: Preparation of Papers for IEEE TRANSACTIONS and JOURNALS}

\corresp{Corresponding author: David Sobrín-Hidalgo (e-mail: dsobh@unileon.es).}

\begin{abstract} %150-250
Building trust between humans and robots has long interested the robotics community. Various studies have aimed to clarify the factors that influence the development of user trust. In Human-Robot Interaction (HRI) environments, a critical aspect of trust development is the robot’s ability to make its behavior understandable. The concept of an eXplainable Autonomous Robot (XAR) addresses this requirement. However, giving a robot self-explanatory abilities is a complex task. Robot behavior includes multiple skills and diverse subsystems. This complexity led to research into a wide range of methods for generating explanations about robot behavior. This paper presents a systematic literature review that analyzes existing strategies for generating explanations in robots and studies the current XAR trends. Results indicate promising advancements in explainability systems. However, these systems are still unable to fully cover the complex behavior of autonomous robots. Furthermore, we also identify a lack of consensus on the theoretical concept of explainability, and the need for a robust methodology to assess explainability methods and tools has been identified.
\end{abstract}


\titlepgskip=-21pt

\maketitle

\section{Introduction}
\label{sec:introduction}
\PARstart{T}{he} integration of autonomous systems, such as robots, into society has advanced alongside technological innovations. Hardware and software developments in the Artificial Intelligence (AI) field have increased the presence of robotic systems among humans. In environments where humans and machines work together, areas such Human-Robot Interaction (HRI) and Human-Robot Collaboration (HRC) have become especially important. The ongoing interaction between humans and robots has created new issues and challenges over the years, drawing interest from the research community.

As a result of this interaction, building trust in robots becomes a key goal in robotics, especially in social and service robotics, where this interaction is most common. Trust is not a new concept in robotics; it has been studied and analyzed over time to propose new methods to increase human trust in robots. For example, robots can be given a humanoid appearance or added expressions and gestures that resemble human behaviors.

Many factors can affect trust building, both positively and negatively. Research in this field shows that the human understanding of a robot's behavior and decisions makes it easier to trust the robot. Explainability addresses this issue. It aims to reduce actions that can appear as a “black box” to the user. Explainability seeks to make robot behavior understandable by providing greater traceability and transparency in its internal processes. Although explainability can apply to different fields, in robotics it is encapsulated in the term eXplainable Autonomous Robot (XAR) \cite{xarSurvey}. According to the authors of the mentioned work, XAR can be interpreted as follows:

\begin{quote}
    "In XAR, the problem is to explain to humans the actions of an autonomous robot operating independently in direct close contact with humans. This can be referred to as goal-driven explainability, or simply as communication."
\end{quote}

The design of an XAR system requires equipping the robot with abilities to explain its actions, decisions, and reasons for its behavior. However, a robot's behavior involves a variety of actuators, sensors, and software subsystems that can impact its tasks. Therefore, understanding the context in the XAR field and the existing methods for generating explanations is essential. This work presents a systematic literature review that aims to identify the methods proposed to generate explanations for autonomous system's behavior. Another review goal is to contextualize these methods by exploring key aspects of explainability, such as application domain, types of explanations generated, and the most common robot abilities used to experiment with explanations.

Before starting the literature review, we conducted a preliminary search to identify similar studies and to build a semi-reference set, as detailed in Section \ref{sub:semireference}. First, the search process to gather other literature reviews on explainability in robotics was conducted using generic search platforms.

Most of the reviews focus on eXplainable Artificial Intelligence (XAI) rather than its application to autonomous robots. Authors of \cite{kumar2023study} analyzed XAI trends, examining the topics and domains of the reviewed articles. This review only analyzes articles from one source (Scopus) and focuses on contextualizing the diverse applications, trends, and methodologies associated with XAI.

The work presented in \cite{vilone2020explainable} proposes a more extensive analysis of XAI trends. The authors study explanations' nature, structure, type, and format. They also analyze some methods used to generate explanations, linking them to the formats in which these explanations are presented to the user. However, the review focuses on XAI and considers articles addressing various AI issues, such as classification, regression, and applications in fields like computer vision. However, the paper did not specifically address robotics as a standalone domain.

On the other hand, the review presented in \cite{wallkotter2021explainable} is more closely related to the robotics field. However, the work primarily focuses on theory and defining key terms, categorizing social cues used for explainability in embodied agents, and organizing evaluation methods.

Finally, authors of \cite{anjomshoae2019explainable} perform an analysis to classify explanations, from application scenarios to the design and evaluation of the explanations themselves.

After the preliminary study of literature reviews on explainability, we identified the need for an analysis focused on the XAR field. Therefore, this review aims to identify and understand the methods proposed by different authors alongside the literature to equip an autonomous robot with self-explanatory capabilities. For this reason, the studies reviewed in this introduction have been used as reference.

This article is structured as follows: Section \ref{sec:methodology} presents the methodology used to guide the literature review process, outlining each phase involved. Section \ref{sec:results} shows the results obtained from each review phase. In Section \ref{sec:discussion}, the results are discussed. Finally, Section \ref{sec:conclusiones} presents the conclusions of the work.

\section{Methodology}
\label{sec:methodology}
This section defines the methodology that guided the proposed systematic review. This methodology is based on the PRISMA guide \cite{moher_prisma_2010} and the recommendations of Kitchenham, Budgen, and Brereton \cite{method_kitchenham_2015}. The type of study conducted is a systematic literature mapping. The general process required for conducting this type of review is illustrated in Figure \ref{fig:mapeoDiagrama}.

\Figure[t!](topskip=0pt, botskip=0pt, midskip=0pt)[width=0.8\linewidth]{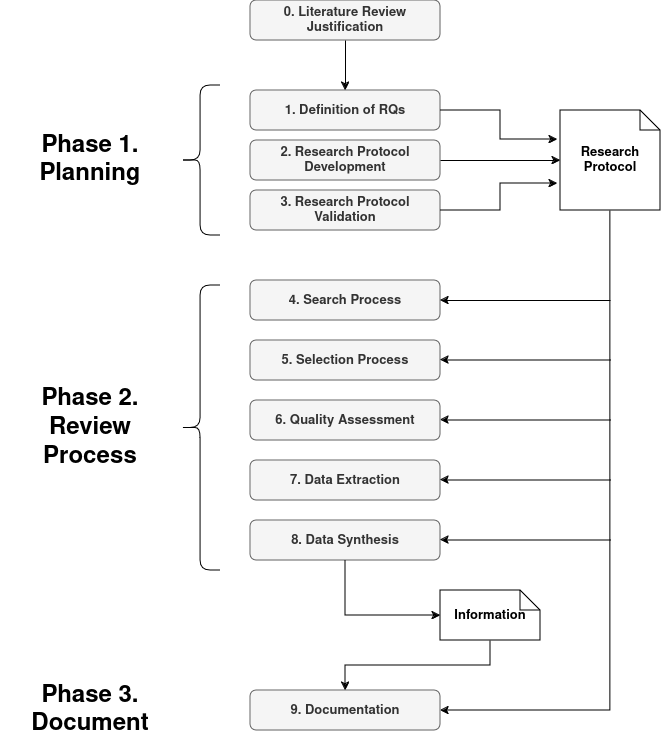}
{\textbf{General structure of a systematic review process. Figure adapted from \cite{method_kitchenham_2015}.}\label{fig:mapeoDiagrama}}

\subsection{Semireference}
\label{sub:semireference}
As a preliminary step to conducting the literature review, we gathered eight semi-reference articles. These articles were selected based on information obtained from the preliminary search outlined in the introduction, previous work by the authors of this review, and recommendations from other researchers with expertise in the field.

The semi-reference set serves several purposes. On one hand, it allows for the extraction of keywords and helps establish the direction of the research. On the other hand, it will serve as a validation method for the search process. This method will be detailed in Section \ref{ssub:searchValidation}. Table \ref{tab:semiref} lists all the papers that form the semi-reference set.

\begin{table}
\caption{\textbf{Semi-reference set.}}
\label{tab:semiref}
\setlength{\tabcolsep}{3pt}
\begin{tabular}{|p{25pt}|p{35pt}|p{165pt}|}
\hline
ID & Reference & Title \\
\hline
\(SR_{1}\) & \cite{Gonz_lez_Santamarta_2023} & Using Large Language Models for Interpreting Autonomous Robots Behaviors\\
\(SR_{2}\) & \cite{han_building_2021} & Building the foundation of explanation generation using  behaviour trees \\
\(SR_{3}\) & \cite{sakai_explainable_2022} & Explainable autonomous robots: a survey and perspective \\
\(SR_{4}\) & \cite{halilovic2023visuo} & Visuo-Textual Explanations of a Robot’s Navigational Choices \\
\(SR_{5}\) & \cite{explainingInTime} & Explaining in Time: Meeting Interactive Standards of Explanation for Robotic Systems\\
\(SR_{6}\) & \cite{fernandez2023accountability} & Accountability and Explainability in Robotics: A Proof of  Concept for ROS
2- And Nav2-Based Mobile Robots \\
\(SR_{7}\) & \cite{sakai_implementation_2023} & Implementation and Evaluation of Algorithms for Realizing Explainable Autonomous Robots \\
\(SR_{8}\) & \cite{borgo_towards_2018} & Towards providing explanations for AI Planner Decisions \\
\hline
\end{tabular}
\end{table}

\subsection{Research Questions}
\label{sub:questions}
The next step in the review involves defining the research questions that will guide the literature analysis. To construct the research questions, we employ the PICOC strategy (\textit{Population, Intervention, Comparison, Outcome, Context}) \cite{schardt_utilization_2007}. This method originated in the Health Sciences field but has been adapted for use in other areas, such as information technology. PICOC aims to make the process of identifying key elements easier. These key elements are necessary for constructing a well-defined research question and identifying keywords for building the search strings.

\begin{table}
\caption{\textbf{PICOC Items}}
\label{tab:picoc}
\setlength{\tabcolsep}{3pt}
\begin{tabular}{|p{75pt}|p{100pt}|p{50pt}|}
\hline
Keywords & Synonyms & Related to \\
\hline
algorithm & answer questions & Outcome\\
                 & answering questions &  \\
                 & framework & \\
                 & system & \\
explainable autonomous robot & XAR & Population \\
                 & behavior explanation &  \\
                 & robot explanation & \\
                 & explanatory capabilities & \\
                 & explain & \\
                 & explainability & \\
                 & explainable & \\
                 & explaining & \\
                 & explanation & \\
                 & explanatory & \\
HRI & human-robot interaction & Context \\
                 & human-robot colaboration &  \\
                 & human-robot partnering & \\
                 & social robots & \\
                 & autonomous agent & \\
                 & autonomoys robot & \\
                 & robot & \\
                 & robot actions & \\
                 & robot behavior & \\
                 & robot experience & \\
                 & robotics & \\
interpretability & interpretable & Population \\
                 & interpreting &  \\
                 & understand & \\
                 & understandable & \\
narrations & verbalization & Comparison \\
                 &visualization &  \\
providing explanation & explanation generation & Intervention \\
                 & generate explanations &  \\
\hline
\end{tabular}
\end{table}

By applying the PICOC strategy, we obtained some terms that guide the research. These terms are presented in Table \ref{tab:picoc} and were obtained by analyzing the works that make up the previously defined semi-reference set. Based on the analysis of the results obtained by applying the PICOC strategy, we define the following Research Question (RQ):

\begin{itemize}
    \item[\textbf{RQ1}] In the field of robotics in general, and HRI specifically, what methods are proposed in the literature for generating explanations?
\end{itemize}

The previously defined question aims to guide the literature review and establish the main goal of gathering some of the current proposals for generating explanations. However, we also defined some secondary research questions (SQ). These questions are listed below:

\begin{itemize}
    \item[\textbf{SQ1}] What is the most commonly used format and type of explanation in the literature?
    \item[\textbf{SQ2}] In which domains is explainability being applied? Which robot skill is most frequently explained?
    \item[\textbf{SQ3}] Are there standardized methods for evaluating the quality of explanations generated by an autonomous system?
\end{itemize}

The complexity of the studied topic requires defining two sets of questions (one primary question and three secondary questions). Although the priority of this research is to answer the primary question (RQ1), we consider the information obtained from the SQs highly valuable for developing the state of the art.

\subsection{Search Planification}
\label{sub:metSearch}
To conduct the search process, we first established the information sources to be used, the restrictions that limit the search results, and the validation strategy needed to assess the results' acceptability. Figure \ref{fig:busquedaDiagrama} illustrates the process we follow.

\Figure[t!](topskip=0pt, botskip=0pt, midskip=0pt)[width=0.9\linewidth]{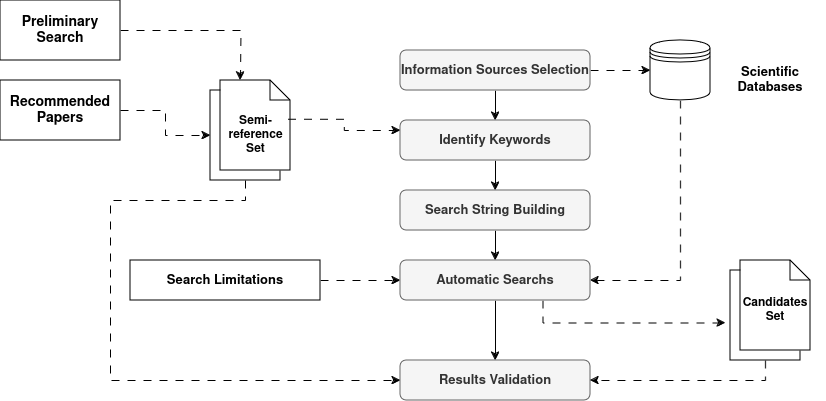}
{\textbf{General structure of search process.}\label{fig:busquedaDiagrama}}

\subsubsection{Information Sources}
\label{ssub:fuentes}
To conduct the search process, we selected a multidisciplinary set of databases to avoid potential information biases. Two of the sources selected are specific to the field of information technology, and the remaining three are general-purpose databases. The selected information sources are listed below:

\begin{enumerate}
    \item ACM Digital Library
    \item IEEE Digital Library
    \item ISI Web of Science
    \item Scopus
    \item Springer Link
    \item Semi-reference\footnote{The items that belong to the semi-reference set and listed in the Table \ref{tab:semiref} from Section \ref{sub:semireference} have been included within their own source of information for consideration during the analysis of results.}
\end{enumerate}

\subsubsection{Search Limitations}
\label{ssub:restricciones}
This section outlines the limitations applied during the search process. These limitations allow for filtering the results we obtained after applying search strings to the selected information sources. These limitations help reduce the number of results based on initial characteristics unrelated to the papers' content (e.g., publication year or language). During this literature review, we selected and applied the following limitations:

\begin{itemize}
    \item Only papers published in 2015 or later are included. This year was selected after reviewing the number of articles published annually. Starting in 2015 (and up to 2018), there is a clear increase in the number of published articles on explainability, as can be observed in the literature review presented in \cite{anjomshoae2019explainable}.
    \item Only papers published in English or Spanish will be selected.
    \item Selected papers must have been published in journals or conferences, excluding other types of literature.
    \item The search will only be performed taking into account the keywords, the title, and the abstract.
    \item Only papers that belong to the Computer Science field are included.
\end{itemize}

\subsubsection{Search Process Validation}
\label{ssub:searchValidation}
Once the search was complete, we gathered an initial set of papers. Verifying that this set is adequate is necessary to ensure the search process was conducted correctly. We applied the criterion proposed in \cite{zhang_validation_2011} to address this validation process. This criterion requires that at least 70\% of the semi-reference papers appear in the search results. We consider the search valid if the established condition is met after applying the mentioned criterion. Otherwise, we need to reformulate the search string to obtain a more accurate set of articles.

\subsection{Search Process}
\label{sub:metSearchProcess}
The search process begins by defining the search strings for each of the previously selected information sources. The build of the search strings used for this systematic mapping was based on elements we identified through the PICOC strategy. We also employed some keywords extracted from the title, abstract, and keywords of the semi-reference articles to build the strings.

After several iterations, three search strings were defined. First, we established a general search string Search String 1. We used this first string as a reference to build the remaining search strings. As mentioned in the limitations shown in Section \ref{ssub:restricciones}, we applied the search strings only to the article title, abstract, and keywords. We chose this approach to avoid an excessive number of irrelevant articles in the search results that do not align with the objectives of this study. The general search string is presented below:

\paragraph{Search String 1 (WoS)} ("XAR" OR "explainable autonomous robots" OR "explainable autonomous robot" OR "explainability") AND ("human-robot" OR "HRI" OR "social robot" OR "robots" OR "robot" OR "robotic" OR "robotics")\newline

The next step consists of defining the remaining search strings for the information sources that require modifications in them. During this process, we considered the nature of the different information sources because some databases are not specific to the field of Computer Science and could require more restrictive search strings. Is important to note that each information source could require a specific format to limit the search to the selected fields (title, abstract, and keywords). The reader can find the search strings in Appendix \ref{app:searchStrings}.

\subsection{Selection Process}
\label{sub:metSelection}
This section describes the sample selection process. The previously defined search process generates a set of \textbf{candidate} papers. The selection process involves applying several filtering stages to reduce this initial set. This filtering process aims to identify papers that are the most relevant to answering the research questions. The process begins after applying the search constraints defined in Section \ref{ssub:restricciones}. The different stages of the selection process are shown in Figure \ref{fig:procesoSelec}.

\Figure[t!](topskip=0pt, botskip=0pt, midskip=0pt)[width=0.9\linewidth]{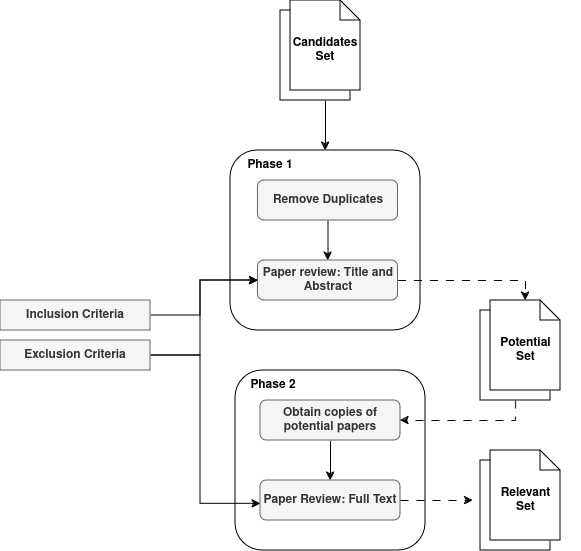}
{\textbf{Structure of the sample selection process.}\label{fig:procesoSelec}}

The first phase of the selection process involves eliminating duplicates and reviewing papers based only on their titles and abstracts. As a result of this step, the candidate papers set is reduced, and we obtain a new set of \textbf{potential} papers. Following this, we applied the second phase of the selection process to this new set of papers. In this final phase, the full text of the articles is read, and those papers that do not meet the established criteria are discarded, generating a set of \textbf{relevant} articles.

We use inclusion and exclusion criteria during the selection process to decide what papers are accepted and what papers are rejected. This criteria is defined in the following section: Section \ref{ssub:inclusion-exclusion}.

\subsubsection{Inclusion and Exclusion Criteria}
\label{ssub:inclusion-exclusion}
The inclusion and exclusion criteria (IC and EC, respectively) used during the sample selection process are defined below. These criteria are also listed in Table \ref{tab:inclusion-exclusion}.

To prevent an ineligible article from being accepted by passing the automatic filters applied in the information sources, criteria EC2 and EC5 have been added. Is important to note that these two criteria are also defined as constraints in Section \ref{ssub:restricciones}.

\begin{table}
\caption{\textbf{Sample inclusion and exclusion criteria}}
\label{tab:inclusion-exclusion}
\setlength{\tabcolsep}{3pt}
\begin{tabular}{|p{25pt}|p{160pt}|p{40pt}|}
\hline
ID & Item & Category\\
\hline
IC1 & The paper is part of the semi-reference set & Inclusion\\
IC2 & The paper implements algorithms or frameworks aimed at generating explanations of the behavior of autonomous systems & \\
EC1 & The main objective of the paper is not explainability for autonomous robots & Exclusion\\
EC2 & The paper is previous to 2015 & \\
EC3 & The paper is a literature review or theoretical analysis & \\
EC4 & The paper is not part of the Computer Science area & \\
EC5 & The paper is not available in English or Spanish & \\
EC6 & Access to the paper is not available & \\
\hline
\end{tabular}
\end{table}

\subsection{Samples Quality Assessment}
\label{sub:evalPlan}
Once the sample selection process is completed, it is necessary to evaluate the quality of the papers within the relevant set. This process serves two purposes: first, it reduces the sample set to be analyzed, and second, it aims to ensure that the studies analyzed demonstrate a certain level of scientific quality.

To conduct the quality evaluation process, we created a checklist based on the proposal by Dybå and Dingsøyr in \cite{DybraQualityEvidences}. The original checklist was adapted to meet the objectives of this study by combining some of the original questions, which aim to assess the general scientific quality of each item, with new questions designed to ensure the sample's relevance to the objectives of this literature review. The control questions (CQ) included in the checklist are presented in Table \ref{tab:listaControl-preguntas}. At the same time, the three possible response types (CA) and their corresponding weights are shown in Table \ref{tab:listaControl-respuestas}.

\begin{table}
\caption{\textbf{Control Quality Questions}}
\label{tab:listaControl-preguntas}
\setlength{\tabcolsep}{3pt}
\begin{tabular}{|p{20pt}|p{155pt}|p{50pt}|}
\hline
ID & Item & Category\\
\hline
CQ1 & Are the paper's goals clearly defined? & Communication\\
CQ2 & Does the paper perform an experiment to evaluate the proposed method? & Rigor\\
CQ3 & Were the data collected related to the research? & Rigor\\
CQ4 & Was the data analysis sufficiently rigorous? & Rigor\\
CQ5 & Does the research explore its applicability to robots (physical or simulated)? & Relevance\\
CQ6 & Does the paper define the method used to generate explanations? & Relevance\\
CQ7 & Does the paper provide the explanations generated during the research? & Relevance\\
CQ8 & Does the paper establish any method for evaluating the explanations generated? & Relevance\\
CQ9 & Is the research conducted in the paper replicated? & Credibility\\
CQ10 & Is there a clear definition in the conclusions? & Credibility\\
\hline
\end{tabular}
\end{table}

\begin{table}
\caption{\textbf{Possible control answer for quality questionary.}}
\label{tab:listaControl-respuestas}
\setlength{\tabcolsep}{3pt}
\begin{tabular}{|p{25pt}|p{160pt}|p{40pt}|}
\hline
ID & Description & Weight\\
\hline
CA1 & Yes & 1.0\\
CA2 & Partially & 0.5\\
CA3 & No & 0.0\\
\hline
\end{tabular}
\end{table}

After applying the checklist, the method used to reduce the sample set consists of establishing a quality threshold based on the score calculated from the answers. Thus, all papers scoring equal to or below the threshold value are rejected. The minimum score a paper must achieve to avoid rejection is 7.0 out of 10.0. This value was set to balance quality and relevance, favoring articles that could provide the necessary information to answer the established research questions.

\subsection{Information Extraction}
\label{sub:extraccionPlan}
We designed a form to conduct the information extraction process. Table \ref{tab:formExtraccion} lists the data that will be extracted to fill in the form. The form includes both general data about the selected papers and data relevant to the defined research questions. We digitally applied the form to each paper of the final set, using the Parsifal tool \footnote{Link to the Parsifal website: https://parsif.al/}.

\begin{table}
\caption{\textbf{Fields of the form used to data extraction of relevant papers}}
\label{tab:formExtraccion}
\setlength{\tabcolsep}{3pt}
\begin{tabular}{|p{100pt}|p{50pt}|p{70pt}|}
\hline
Item & Type & Possible Values\footnote{Column "Possible Values" shows only some examples of the values to make the table easy to read.}\\
\hline
Title & \textit{String} & n/a\\
Identifier & \textit{int} & n/a\\
Reference & \textit{String} & n/a\\
Publication Date & \textit{int} & \(\geq\) 2015\\
Suggested Method  & \textit{String} & n/a\\
Do the authors ask questions to the system? & \textit{Boolean} & Yes/No\\
Application Domain & Multiple Answer & Industrial\\
                  &                    & Medicine \\
                  &                    & Social \\
                  &                    & Services \\
                  &                    & Agriculture \\
                  &                    & Military \\
                  &                    & Space \\
                  &                    & Educational \\
                  &                    & Not defined \\
Robot's skill & Multiple Answer & Perception\\
                  &                    & Navigation \\
                  &                    & Manipulation \\
                  &                    & Decision-making process \\
                  &                    & Interaction \\
Does the system provide real-time explanations? & \textit{Boolean} & Yes/No\\
Explanation Type & Single answer & Multimodal\\
                    &                   & Textual\\ 
                    &                   & Visual\\
                    &                  & Other\\
Variables of interest & \textit{String} & E.g.: Time, level of detail, duration...\\
Environment & Multiple answer & Simulation\\
                                   &                     & Real Word\\
                                   &                     & Both\\
Do the authors propose a method for the evaluation of explanations? & \textit{Boolean} & Yes/No\\
\hline
\end{tabular}
\end{table}

A popular trend in the study of explainability for autonomous systems involves framing explanations as responses to specific user questions. The nature of these questions is a point of interest for this research. However, given the potentially wide variety of questions, we only took into account the number of articles that generate explanations based on a question-answer system rather than each specific question.

On the other hand, the defined form includes two elements (application domain and robot skill). Those elements provide the information needed to develop an overview of explainability studies’ experimentation trends. However, both the application domain and robot skill are broad concepts. Therefore, to simplify the information extraction and synthesis process, we have established predefined groups for classifying application domains (Section \ref{ssub:dominios}) and robot skills (Section \ref{ssub:skills}). This way, information extraction will focus only on classifying each paper’s experimentation within these predefined groups.

\subsubsection{Classification of Application Domains}
\label{ssub:dominios}
The following list shows some of the possible robotic application domains considered for the result extraction stage of this review. Is important to note that this list may include more domains than those represented in the studied articles.

\begin{enumerate}
    \item \textbf{Industrial Robotics}: This includes robots operating in factories or warehouses, performing tasks such as assembling, packaging, inventory management, and sorting.
    \item \textbf{Healthcare Robotics}: robots designed for surgical tasks, patient assistance in rehabilitation...
    \item \textbf{Social Robotics}: robots that interact and collaborate with humans in social settings, as well as domestic robots.
    \item \textbf{Services Robotics}\footnote{Most often, social robotics and service robotics are part of the same domain. However, in this study, they have been divided into two separate domains to provide a clearer distinction between the types of tasks performed.}: This includes robots that perform services for humans in environments with minimal interaction. Examples in this domain include delivery drones, autonomous vehicles, and commercial robots.
    \item \textbf{Agricultural Robotics}: This domain includes robots responsible for farming and harvesting tasks, as well as autonomous systems for livestock management.
    \item \textbf{Military Robotics}: includes reconnaissance and combat robots, troop transport robots, robots used for bomb deactivation, and similar applications in military and defense contexts.
    \item \textbf{Spacial Robotics}: includes robots used for exploration, mapping, or monitoring tasks in space. It involves rovers, satellites, and other space-based autonomous systems.
    \item \textbf{Educational Robotics}: platforms used for teaching.
\end{enumerate}

In some cases, due to the complexity of the explainability field, the experiments in the papers may not be encapsulated in a specific domain but rather in the robot's capabilities. This means that the research could apply to multiple domains. For articles affected by this situation, the label "Not defined" has been established as a possible value for the application domain.

To establish the application domain in which each of the analyzed papers is developed, we have evaluated both the context of the proposed research and the nature of the experiments conducted.

\subsubsection{Skills del robot}
\label{ssub:skills}
Different robots can have various capabilities or skills to interact with the environment. The possible skills that an autonomous system can integrate are highly varied. To facilitate the information extraction process and condense the results, we have defined the following skill groups, which will be taken into account when applying the information extraction form:

\begin{enumerate}
    \item \textbf{Perception}: skills related to sensors, object recognition, environment mapping, etc.
    \item \textbf{Navigation}: system capabilities that allow it to move through an environment (walking, flying, underwater movement...)
    \item \textbf{Manipulation}: skills derived from the use of robotic arms
    \item \textbf{Decision-making skills}: skills related to planning, reasoning, problem-solving, error detection and recovery, etc.
    \item \textbf{Interaction}: skills that enable the system to interact and communicate with humans. Examples are natural language processing capabilities or human-robot dialog.
\end{enumerate}

Despite the defined list, some of the tasks that a robot can perform may require the use of different skills classified into various groups simultaneously. For this reason, the element of the information extraction form corresponding to the robot's skills is kept as a multiple-choice response.

\section{Results}
\label{sec:results}
Throughout the previous section, we established the methodology that will guide the various stages of the systematic review. This section presents the results obtained in each phase outlined in the methodology.

\subsection{Search Process Results}
\label{sub:serachRes}
The search process's results are detailed below. The data presented in this section do not include the papers that belong to the semi-reference set, as these were pre-selected and later used to validate the search process itself.

First, an initial search was conducted without applying the restrictions defined in Section \ref{ssub:restricciones}. This search yielded a total of 803 results. The distribution of the obtained results is shown in Figure \ref{fig:busquedaInicial}.

\Figure[t!](topskip=0pt, botskip=0pt, midskip=0pt)[width=0.9\linewidth]
{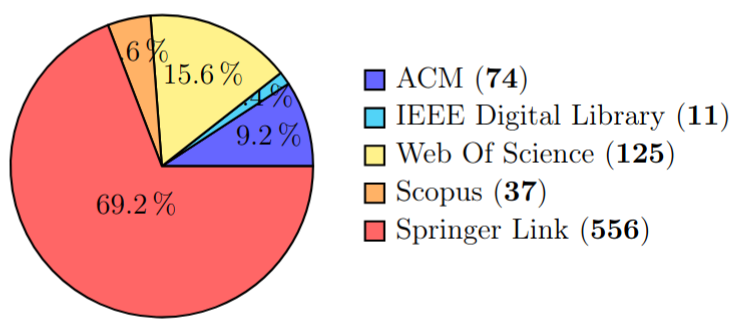}
{\textbf{Initial search process output distribution. No search limitation applied.}\label{fig:busquedaInicial}}

After the initial search, the sample set was reduced from the original total (803 samples) to a new set of 129 papers. This new filtering step was carried out considering the search restrictions defined earlier and using the filtering tools provided by the different information sources. The distribution of the articles in this new sample set is shown in Figure \ref{fig:busquedaRestringida}.

\Figure[t!](topskip=0pt, botskip=0pt, midskip=0pt)[width=0.9\linewidth]
{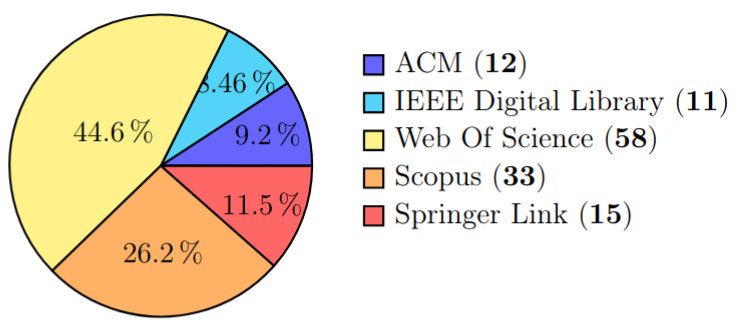}
{\textbf{Initial search process output distribution. Information source limitations applied.}\label{fig:busquedaRestringida}}

Once the search was completed, we applied the validation method for the resulting set, as defined in Section \ref{ssub:searchValidation}. The search process was considered valid since the sample set contains 6 of the 8 articles from the reference set, which corresponds to approximately 70\%. At this point, the papers from the semi-reference set were added to the new set, raising the total number of samples to 137.

\subsection{Selection Process Results}
\label{sub:seleccionRes}
The selection process consists of two main phases, as shown in Figure \ref{fig:procesoSelec}. The inclusion and exclusion criteria established in Section \ref{ssub:inclusion-exclusion} were applied during both phases. The results obtained in each defined phase are detailed in the following sections.

\subsubsection{Phase 1 of the Selection Process}
\label{ssub:selFase1}
We applied the first phase of the sample selection to the group of candidate papers obtained as the search process's output. A total of 26 duplicates were automatically removed, reducing the candidate set from 137 to 111 articles. After removing the duplicates, we performed a filtering process based on reading the title and the abstract. This filtering allowed the exclusion of 73 articles, resulting in a set of 38 potential papers.

\subsubsection{Phase 2 of the Selection Process}
\label{ssub:Fase2}
In this second phase, we reviewed the set of potential papers obtained from Phase 1 by reading the full text of each paper. During this phase, we labeled one paper that had passed the automatic filter in the previous phase as a duplicate. Additionally, 8 articles were rejected after reading the full text. Once again, all articles from the semi-reference set were accepted according to the inclusion criterion IC1. This phase produced a new set of 29 relevant papers as its output.

\subsection{Evaluation Process Results}
\label{sub:evaluacionRes}
During the evaluation phase, we applied the form to each paper belonging to the relevant paper set. After applying the questionnaire and obtaining the score for each paper, all articles with a score lower than 7.0 points were discarded. A total of 7 articles were discarded in that way. Among the rejected articles, the lowest score obtained was 3.5, while the remaining articles had scores between 5.0 and 7.0 points. The articles accepted after this stage form the \textbf{set of final papers}. This final set consisted of 22 items, with the highest score among the items in the set being 9.5 points.

In this stage, we applied the evaluation process to the semi-reference set. After this stage of the review, two articles from the semi-reference set (\(SR_{3}\) and \(SR_{8}\)) were discarded for not reaching the minimum expected score, with scores of 5.0 and 3.5, respectively. It is worth noting that the semi-reference article \(SR_{3}\) corresponds to a literature review. While this article was useful for establishing keywords and constructing search strings related to the research field, it is not relevant for addressing the research questions defined. Discarding this article through the evaluation questionnaire indicates that the relevance criteria were correctly established. Table \ref{tab:definitivo} lists the finally accepted papers. The table also shows the publication date and the paper's score.

\begin{table}[t]
\caption{\textbf{Items of the final set of papers.}}
\label{tab:definitivo}
\setlength{\tabcolsep}{3pt}
\begin{tabular}{|p{20pt}|p{20pt}|p{130pt}|p{20pt}|p{20pt}|}
\hline
ID & Ref. & Title & Date & Score \\
\hline
\(AD_{1}\) & \cite{halilovic2023influence} & The Influence of a Robot’s Personality on Real-Time Explanations of Its Navigation & 2023 & 8.0 \\
\(AD_{2}\) & \cite{han2023communicating} & Communicating Missing Causal Information to Explain a Robot’s Past Behavior & 2023 & 8.0 \\
\(AD_{3}\) & \cite{rosenthal2022impact} & The Impact of Route Descriptions on Human Expectations for Robot Navigation & 2022 & 9.0 \\
\(AD_{4}\) & \cite{sakai_implementation_2023} & Implementation and Evaluation of Algorithms for Realizing ExplainableAutonomous Robots & 2023 & 9.5 \\
\(AD_{5}\) & \cite{explainingInTime} & Explaining in Time: Meeting Interactive Standards of Explanation forRobotic Systems & 2021 & 7.5 \\
\(AD_{6}\) & \cite{cruz2023explainable} & Explainable robotic systems: understanding goal-driven actions in areinforcement learning scenario & 2021 & 7.5 \\
\(AD_{7}\) & \cite{hu2023explainable} & Explainable autonomous robots in continuous state space based ongraph-structured world model & 2023 & 9.0 \\
\(AD_{8}\) & \cite{BOGATARKAN_ERDEM_2020} & Explanation Generation for Multi-Modal Multi-Agent Path Finding withOptimal Resource Utilization using Answer Set Programming & 2020 & 7.5 \\
\(AD_{9}\) & \cite{mualla2022quest} & The quest of parsimonious XAI: A human-agent architecture forexplanation formulation & 2021 & 7.5 \\
\(AD_{10}\) & \cite{zakershahrak2020online} & Online Explanation Generation for Planning Tasks in Human-Robot Teaming & 2020 & 8.0 \\
\(AD_{11}\) & \cite{mota2021answer} & Answer me this: Constructing Disambiguation Queries for Explanation Generation in Robotics & 2021 & 8.0 \\
\(AD_{12}\) & \cite{Gonz_lez_Santamarta_2023} & Using Large Language Models for Interpreting Autonomous Robots Behaviors & 2023 & 8.0 \\
\(AD_{13}\) & \cite{gong2018behavior} & Behavior Explanation as Intention Signaling in Human-Robot Teaming & 2018 & 8.5 \\
\(AD_{14}\) & \cite{han_building_2021} & Building the Foundation of Robot Explanation Generation Using Behavior Trees & 2021 & 8.0 \\
\(AD_{15}\) & \cite{schroeter2022introspectionbasedexplainablereinforcementlearning} & Introspection-based Explainable Reinforcement Learning in Episodic and Non-episodic Scenarios & 2022 & 7.5 \\
\(AD_{16}\) & \cite{gao2020joint} & Joint Mind Modeling for Explanation Generation in Complex Human-Robot Collaborative Tasks & 2020 & 8.5 \\
\(AD_{17}\) & \cite{brandao2021towards} & Towards Providing Explanations for Robot Motion Planning & 2021 & 9.0 \\
\(AD_{18}\) & \cite{chen2020towards} & Towards transparent robotic planning via contrastive explanations & 2020 & 8.0 \\
\(AD_{19}\) & \cite{olivares2021knowledge} & Knowledge representation for explainability in collaborative robotics and adaptation & 2021 & 7.5 \\
\(AD_{20}\) & \cite{brandao2021explaining} & Explaining Path Plan Optimality: Fast Explanation Methods for Navigation Meshes Using Full and Incremental Inverse Optimization & 2021 & 9.0 \\
\(AD_{21}\) & \cite{halilovic2023visuo} & Visuo-Textual Explanations of a Robot's Navigational Choices & 2023 & 8.0 \\
\(AD_{22}\) & \cite{fernandez2023accountability} & Accountability and Explainability in Robotics: A Proof of Concept for ROS 2-And Nav2-Based Mobile Robots & 2023 & 7.5 \\
\hline
\end{tabular}
\end{table}

Finally, we analyzed the publication trend of the articles after finishing the evaluation process and the subsequent building of the final set of papers. Figure \ref{fig:tendencia} shows a graph reflecting this trend. The graph shows that the highest number of articles focused on explainability techniques for autonomous systems was published in 2021. Although the number of articles decreased significantly in 2022, likely due to the COVID-19 pandemic over the previous two years. The number of publications increased again in 2023, indicating that the study of explainability in such systems remains a topic of high interest for the scientific community.

\Figure[h!](topskip=0pt, botskip=0pt, midskip=0pt)[width=0.9\linewidth]
{TendenciaArticulosAno.PNG}
{\textbf{Number of relevant papers per year.}\label{fig:tendencia}}

\subsection{Information Extraction Process Results}
\label{sub:extracionRes}
This section presents the results obtained in the current literature review. The numerical results related to building the final sample set are shown in the PRISMA diagram (Preferred Reporting Items for Systematic Reviews and Meta-Analysis), which can be found in Figure \ref{fig:prisma}.

\Figure[h!](topskip=0pt, botskip=0pt, midskip=0pt)[width=0.9\linewidth]
{DiagramaPrisma.PNG}
{\textbf{PRISMA diagram for the identification, selection, and quality evaluation processes.}\label{fig:prisma}}

This section also gathers the information extracted from the final article set. The data was collected by applying the extraction questionnaire defined in Section \ref{sub:extraccionPlan}. The extracted data is presented in three different tables. Table \ref{tab:general} contains general information related to the scope and experimentation of the analyzed studies. It is important to note that one of the 22 papers focuses its research on two application domains simultaneously: service robotics and industrial robotics. Secondly, Table \ref{tab:expli} presents a series of boolean data regarding the use of explanations in the studies reviewed. Finally, Table \ref{tab:methods} presents information related to the methods proposed in the different studies. The table includes the method proposed by the authors, the type of explanation generated, and the format in which the explanation is presented.

\begin{table}[h!]
\caption{\textbf{Data related to the nature of the research conducted in each selected article.}}
\label{tab:general}
\setlength{\tabcolsep}{3pt}
\begin{tabular}{|p{20pt}|p{60pt}|p{50pt}|p{85pt}|}
\hline
ID & Application Domain & Environment & Robot's Skills\\
\hline
 \(AD_{1}\) & Social Robotics & Simulation & Navigation\\
 \(AD_{2}\) & Industrial Robotics & Real World & Manipulation, Navigation\\
 \(AD_{3}\) & Not Defined & Simulation & Navigation\\
 \(AD_{4}\) & Social Robotics& Simulation & Decision-making\\
 \(AD_{5}\) & Social Robotics& Simulation & Manipulation, Navigation\\
 \(AD_{6}\) & Social Robotics & Both & Manipulation, Navigation, Perception\\
 \(AD_{7}\) & Not Defined & Simulation & Navigation, Decision-making\\
 \(AD_{8}\) & Industrial Robotics & Simulation & Decision-making\\
 \(AD_{9}\) & Services Robotics & Simulation & Navigation\\
 \(AD_{10}\) & Space Robotics & Simulation & Decision-making\\
 \(AD_{11}\) & Social Robotics & Simulation & Manipulation, Perception\\
 \(AD_{12}\) & Social Robotics & Real World & Perception, Navigation, Manipulation, Decision-making, Interaction\\
 \(AD_{13}\) & Social Robotics & Simulation & Navigation\\
 \(AD_{14}\) & Industrial and Services Robotics & Simulation & Manipulation, Navigation\\
 \(AD_{15}\) & Services Robotics & Simulation & Navigation\\
 \(AD_{16}\) & Social Robotics & Simulation & Manipulation, Navigation\\
 \(AD_{17}\) & Social Robotics & Simulation & Navigation, Decision-making\\
 \(AD_{18}\) & Not Defined & Simulation & Decision-making\\
 \(AD_{19}\) & Social Robotics & Real World & Manipulation\\
 \(AD_{20}\) & Not Defined & Simulation & Navigation, Decision-making\\
 \(AD_{21}\) & Social Robotics & Simulation & Navigation\\
 \(AD_{22}\) & Social Robotics & Simulation & Navigation\\
\hline
\end{tabular}
\end{table}

\begin{table}[t]
\caption{\textbf{Answers to questions about the methods of explanation proposed in the articles analyzed.}}
\label{tab:expli}
\setlength{\tabcolsep}{3pt}
\begin{tabular}{|p{20pt}|p{65pt}|p{60pt}|p{70pt}|}
\hline
ID & Do the authors ask questions to
the system? & Does the system provide real-
time explanations? & Do the authors propose a
method for the evaluation of
explanations?\\
\hline
\(AD_{1}\) & False & True & False \\
\(AD_{2}\) & False & False & True \\
\(AD_{3}\) & False & False & True \\
\(AD_{4}\) & True & False & True \\
\(AD_{5}\) & True & True & False \\
\(AD_{6}\) & True & True & False \\
\(AD_{7}\) & True & False & True \\
\(AD_{8}\) & True & False & False \\
\(AD_{9}\) & True & True & True \\
\(AD_{10}\) & False & True & True \\
\(AD_{11}\) & True & True & False \\
\(AD_{12}\) & True & False & False \\
\(AD_{13}\) & True & True & True \\
\(AD_{14}\) & True & True & False \\
\(AD_{15}\) & True & False & False \\
\(AD_{16}\) & False & False & True \\
\(AD_{17}\) & True & True & True \\
\(AD_{18}\) & False & False & True \\
\(AD_{19}\) & True & True & False \\
\(AD_{20}\) & True & False & True \\
\(AD_{21}\) & True & True & False \\
\(AD_{22}\) & True & False & False \\
\hline
\end{tabular}
\end{table}

\begin{table}[t]
\caption{\textbf{Data on the methods of explanation proposed in the articles analyzed.}}
\label{tab:methods}
\setlength{\tabcolsep}{3pt}
\begin{tabular}{|p{20pt}|p{120pt}|p{40pt}|p{40pt}|}
\hline
ID & Method Description & Explanation Type & Explanation Format \\
\hline
\(AD_{1}\) & HiXRoN, a framework for generating explanations in Navigation tasks, based on environment ontologies and events that occur during the behavior & Descriptive & Multimodal \\
\(AD_{2}\) & Replay-Project-Say, a method that replays the robot's behavior to help the user to understand what occurred & Causal & Visual \\
\(AD_{3}\) & Algorithm based in path and map analysis & Descriptive & Textual \\
\(AD_{4}\) & Algorithm based on the comparison between robot's world model and user's estimated world model & Causal & Visual \\
\(AD_{5}\) & Algorithm based on VEL (Violation Enumeration Language) of temporal logic & Descriptive & Textual \\
\(AD_{6}\) & A method based on analyzing the probability of achieving a goal using Q-Values. It includes three approaches: Memory-based explainable RL, Learning-based, and Introspection-based. & Descriptive & Textual \\
\(AD_{7}\) & Explanation framework based on graph-structured world model & Descriptive & Visual \\
\(AD_{8}\) & Query-based explanation generator algorithm using weighted weak constraints & Descriptive & Textual \\
\(AD_{9}\) & HAExA, explainability architecture for autonomous robots based on beliefs and intention changes & Contrastive & Textual \\
\(AD_{10}\) & Online Explanation Generation (OEG) - as Model Reconciliation. Comparison between robot's model and human's model & Causal & Numeric \\
\(AD_{11}\) & An architecture that traces the evolution of beliefs and aims to reduce ambiguity in human queries & Descriptive & Textual \\
\(AD_{12}\) & Analysis of ROS logs through LLMs & Descriptive & Textual \\
\(AD_{13}\) & Intention signaling through robot's plan analysis & Descriptive & Textual \\
\(AD_{14}\) & Utilization of BTs to explore the robot's behavior & Causal & Textual \\
\(AD_{15}\) & Introspection-based approach using QValues & Causal & Numeric \\
\(AD_{16}\) & Framework based on the human's mental model inference from the robot's model & Causal & Textual \\
\(AD_{17}\) & Two methods to explain planning: one based on optimization and the other based on sampling & Contrastive & Textual \\
\(AD_{18}\) & Explanation of planning from Markov Decision Process (MDP) & Contrastive & Textual \\
\(AD_{19}\) & ARE-OCRA explanation generation algorithm. Provides answers to user questions using a knowledge base built with ontologies & Causal & Textual \\
\(AD_{20}\) & Method that generates explanations from changes in a navmesh graph, the objective is to know the most optimal path & Contrastive & Multimodal \\
\(AD_{21}\) & Combination of LIME, Qualitative Spatial Reasoning (QSR), and ontologies & Causal & Multimodal \\
\(AD_{22}\) & ROS Logs analysis & Causal & Textual \\
\hline
\end{tabular}
\end{table}

\section{Discussion}
\label{sec:discussion}
This section analyzes the results presented in Section \ref{sec:results}. The data obtained and gathered in previous tables are presented here in different charts to make it easier to analyze how often the data of interest appear in the final set of analyzed articles.

First, we will focus the analysis on the general approach of the studies considered. There is a clear trend in both the application domain of the experiments and the environments used. Figure \ref{fig:frecDominio} shows a bar chart that presents the number of papers that focus on each defined domain\footnote{Although a total of 22 articles were analyzed, the number of domains shown in the figure is 23 because one article conducted experiments in two different application domains.}. Results show that social robotics emerges as the most popular field in explainability research. This is an expected result, given the impact of explainability on HRI, a concept closely related to social robotics.

On the other hand, only one article addressed explainability applied to rovers and their use in the space domain. The second-largest group of papers is labeled "Not Defined"; the total number of articles that did not focus their experimentation on a specific domain amounts to four, well below the results obtained for social robotics.

\Figure[t!](topskip=0pt, botskip=0pt, midskip=0pt)[width=0.9\linewidth]
{FrecuenciaDominio.PNG}
{\textbf{Frequency of the different application domains throughout the analyzed articles.}\label{fig:frecDominio}}

Regarding the environment used for experimentation, we have found that simulations are much more commonly used than physical robots, as shown in the pie chart in Figure \ref{fig:Distribucionentornos}. Space and hardware limitations may contribute to the popularity of simulators. It is also important to consider that the date of the papers reviewed coincides with the COVID-19 pandemic, an event that may have impacted the use of simulators due to the in-person restrictions imposed by the pandemic.

\Figure[t!](topskip=0pt, botskip=0pt, midskip=0pt)[width=0.9\linewidth]
{DistribucionEntornos.PNG}
{\textbf{Distribution of the common environments used for the experimentation carried out in the analyzed papers.}\label{fig:Distribucionentornos}}

The experimentation conducted in the reviewed articles was limited to a predefined set of robot skills. Giving self-explanation capabilities to an autonomous system is a complex challenge. To address this issue, we restrict the problem to a few skills to simplify the method's analysis. Figure \ref{fig:frecSkill} shows the frequency of the five groups that classify the different robot skills. 

Results show that Navigation is the most frequently used skill group in explainability experiments. Navigation is a necessary skill for addressing a wide range of tasks in robotics, so it is expected that this skill will be used in various application domains. Another reason for the large number of articles focused on this skill is the existence of many use cases where the application of explanation generation methods is of great interest. Some examples of these use cases are route replanning, person tracking, collision management, etc.

On the other hand, although still much lower than Navigation, Manipulation and Decision-making skills also have a significant presence. It can be concluded that there is a strong relationship between the application domain and the skills involved in the experimentation, as Navigation, Manipulation, and Planning (included in the Decision-making group) are important in both industrial and service/social robotics. Additionally, exploring explainability from planners or Decision-making approaches also seems promising. Focusing explainability on the robot's Decision-making process may allow studying its behavior at a higher level of abstraction, directly analyzing the sequence of actions or tasks the robot performs during the execution of its behavior.

It is important to mention that during the analysis of the frequency of occurrence of different skills, it was identified several times that a single article conducts experimentation involving more than one skill. This is why the total frequency count reaches 36 mentions, even though the set of articles studied consists of only 22 samples.

\Figure[t!](topskip=0pt, botskip=0pt, midskip=0pt)[width=0.9\linewidth]
{FrecuenciaSkill.PNG}
{\textbf{Frequency of skills along the papers analyzed.}\label{fig:frecSkill}}

Regarding the analysis of the methods proposed in the literature, we constructed a general overview of the trends followed by the reviewed research. To do this, we defined three questions with True or False answers during the information extraction process. The papers analyzed had to give responses to each of these questions. The results are presented in Figure \ref{fig:respuestasbool}. After analyzing these results, we can conclude that most papers propose generating explanations as a response to a user question asked to the autonomous system or to the framework responsible for providing explainability to the robot. That is, the user’s question triggers and guides the generation of the explanation. Only 6 of the 22 articles do not use this question-based approach.

\Figure[t!](topskip=0pt, botskip=0pt, midskip=0pt)[width=0.9\linewidth]
{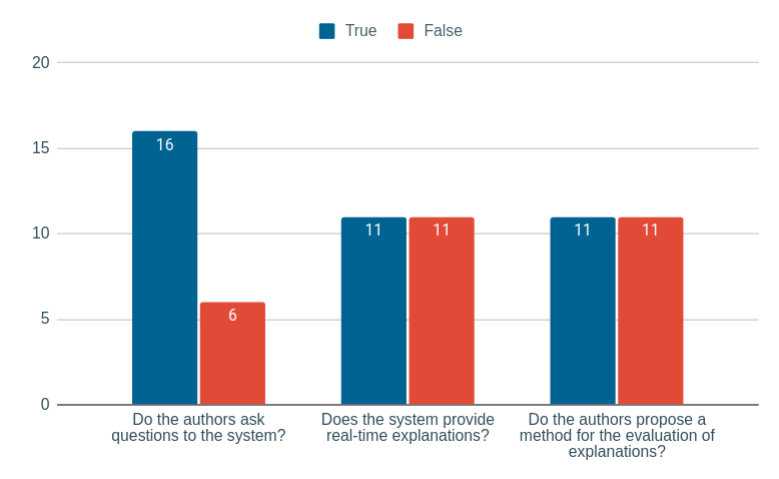}
{\textbf{Trends of different approaches of explainability systems.}\label{fig:respuestasbool}}

There is no clear trend regarding real-time explanation generation, with half of the articles generating this type of explanation. However, we can establish a relationship between real-time explanations and the type of explanation. In general, causal explanations that aim to identify the origin of a specific behavior are provided once that behavior has concluded. On the other hand, of the 11 articles that do not provide real-time explanations, 4 papers offer explanations before the behavior occurs. The goal of these methods is to predict the robot's behavior, as well as the user’s expectations about that behavior.

Once the explanation is generated, it is important to assess its validity for the user. The data analysis indicates that only half of the articles propose a method for evaluating the explanations. The trend among evaluation methods involves defining metrics and conducting user studies. Some of the most commonly used metrics are listed below:

\begin{itemize}
    \item \textbf{Metrics to measure user satisfaction:} Credibility, brevity, understandability, reliability, overwhelming and predictability.
    \item \textbf{Metrics to measure explanation generation}: Response time, number of words per explanation, number of visual objects per explanation...
    \item \textbf{Metrics to measure the adequacy of the explanation}: Accuracy, similarity, effectiveness, efficiency.
\end{itemize}

One of the goals that motivates this literature review is to understand the nature of the explanations provided by the different proposed methods. Figure \ref{fig:tipoFormato} shows the frequency of the different explanations found in the reviewed literature. The figure links the type of the explanation with its format.

\Figure[t!](topskip=0pt, botskip=0pt, midskip=0pt)[width=0.9\linewidth]
{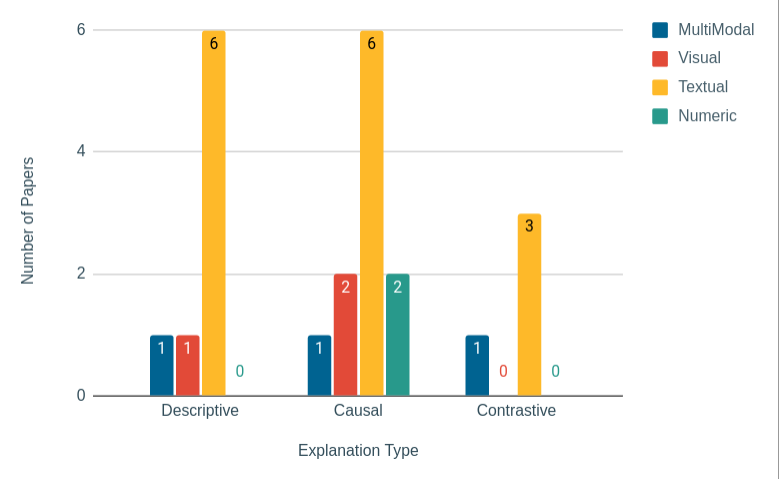}
{\textbf{Relationship between type and format of explanation, and frequency of appearance throughout the articles analyzed.}\label{fig:tipoFormato}}

The graph reveals a higher popularity of text-based explanations, which are closely associated with descriptive explanations. Nearly all the explanations analyzed are presented in a textual format. Causal explanations are the second most popular type of explanation. However, for this category, the format used to present the explanation is more varied. Finally, the results for contrastive explanations, which involve comparing two situations, indicate that they are typically presented in text format. Although visual and numerical formats can be used for this type of explanation, these explanations do not seem to rely on a single format but rather a combination of several, in which case they are classified as a multi-modal format.

Finally, the methods proposed in the reviewed literature will be discussed. Each article studied proposes a method, framework, or algorithm for generating explanations. These methods were previously listed in Table \ref{tab:methods}. Since these explanation-generation mechanisms are not standardized, we have synthesized the information to highlight common ideas and approaches for designing explanation-generation systems. These proposals are discussed below:

\begin{itemize}
    \item \textbf{Probability-based methods} that typically use Q-values to estimate the likelihood of successfully achieving a goal. A decision is made based on the probability of success, so a higher probability represents the cause of the behavior.
    % Métodos introspectivos?
    \item \textbf{Methods based on the analysis of internal data} of the system. That data is extracted from the “logs”, the “behavior trees”, and the map...
    % Métodos generativos?
    \item \textbf{Data inference methods} that use algorithms and techniques to calculate new values from the system's internal data. Then, the method uses the inferred values to construct explanations.
    \item \textbf{Model Reconciliation methods} that aim to study the differences between the robot's model and the model estimated by the human. Understanding these differences makes it possible to generate explanations that aim to enhance mutual understanding between humans and robots.
    \item \textbf{Knowledge analysis methods} that generate explanations by examining the system’s knowledge bases and ontologies.
    \item \textbf{Methods that analyze changes in intentions and beliefs}. These methods generate explanations about the system’s decision-making process.  
\end{itemize}

It is important to mention that the methods presented here do not represent the entirety of the literature. This study aims to highlight the main trends by analyzing a defined set of articles. Numerous methods and proposals cannot be clearly and precisely assigned to a specific group. For example, the paper \(AD_{2}\) proposes a system to replay the robot's behavior and presents this replay as an explanation. Another example is the work presented in \cite{rosenthal2016verbalization}, which was not included in the literature review but is worth mentioning. In this work, the robot’s verbalization is used to improve understanding between the machine and the human. Based on the analysis, we can affirm that the proposed methods are intrinsically linked to the conception of the explanation concept, which highlights the importance of establishing standardized concepts and definitions.

\section{Conclusion}
\label{sec:conclusiones}
We conduct a systematic literature review to identify and analyze different methods proposed to equip autonomous systems with self-explanatory capabilities. The study involved a search phase, sample selection, and quality assessment, followed by a data extraction process. From an initial set of 803 samples, the filtering phases resulted in a final set of 22 articles. These were analyzed to provide an overview of recent trends in the explainability field, with a technical focus emphasizing experimentation. The analysis performed, and the results obtained allow us to answer the research question defined during the planning of this study. 

\paragraph{\textbf{RQ1} In the field of robotics in general, and HRI specifically, what methods are proposed in the literature for generating explanations?} There is a wide range of methods and techniques for generating explanations. However, some of the most commonly used methods in the literature analyze the system's internal data and generate explanations based on the study of the system's model. On the other hand, a common method to address explanation generation is the dialog-based system, where the explanation is presented as an answer to a "Why" type question. 

\paragraph{\textbf{SQ1} What is the most commonly used format and type of explanation in the literature?} Causal and descriptive explanations have proven to be more common than contrastive explanations. On the other hand, regardless of the type of explanation, the most commonly used format for providing explanations is text-based.

\paragraph{{SQ2} In which domains is explainability being applied? Which robot skill is most frequently explained?} Explainability has shown a greater impact and increased interest from the research community in the social robotics domain. Regarding the robot's capabilities, the experimentation conducted in the studied articles tends to focus on navigation tasks.

\paragraph{\textbf{SQ3} Are there standardized methods for evaluating the quality of explanations generated by an autonomous system?} Although there are methods that evaluate the explanations generated by the proposed approaches, half of the analyzed articles do not assess their quality. That result shows a lack of a common methodology that defines the necessary methods and metrics for evaluating an explanation. \newline

In summary, this literature review has highlighted the complexity and breadth of the concept of explainability in the robotics field. Based on the results obtained, the following conclusions have been drawn:

\begin{itemize}
    \item Lack of agreement or standardization of the concept of explanation and the need to establish a classification of explanations.
    \item There is a need to establish a methodology and techniques or mechanisms to evaluate a generated explanation accurately. In other words, it should be possible to determine how adequate an explanation an autonomous system provides is.
    \item The diversity along the explanation generation methods studied reveals a range of promising approaches. However, it is still possible to improve and achieve the proposed goals within the field of explainability.
\end{itemize}

\appendices
% If you have multiple appendices, use the $\backslash$appendices command below. If you have only one appendix, use
% $\backslash$appendix[Appendix Title]

\section{Search Strings}
\label{app:searchStrings}
This appendix includes the search strings that were constructed from the initial search string to address the needs of the different information sources used.

\paragraph{Search String 2 (ACM)} {[[[Title: "xar"] OR [Title: "explainable autonomous robots"] OR [Title: "explainable autonomous robot"] OR [Title: "explainability"] OR [Title: "explaining"]] AND [[Title: "human-robot"] OR [Title: "hri"] OR [Title: "social robot"] OR [Title: "robots"] OR [Title: "robot"] OR [Title: "robotics"]]] OR [[[Abstract: "xar"] OR [Abstract: "explainable autonomous robots"] OR [Abstract: "explainable autonomous robot"] OR [Abstract: "explainability"] OR [Abstract: "explaining"]] AND [[Abstract: "human-robot"] OR [Abstract: "hri"] OR [Abstract: "social robot"] OR [Abstract: "robots"] OR [Abstract: "robot"] OR [Abstract: "robotics"]]] OR [[[Keywords: "xar"] OR [Keywords: "explainable autonomous robots"] OR [Keywords: "explainable autonomous robot"] OR [Keywords: "explainability" or "explaining"]] AND [[Keywords: "human-robot"] OR [Keywords: "hri"] OR [Keywords: "social robot"] OR [Keywords: "robots"] OR [Keywords: "robot"] OR [Keywords: "robotics"]]]}

\paragraph{Search String 3 (IEEE)} (((("Publication Title": "XAR" OR "explainable autonomous robot*"OR "explainability" OR "interpretability" OR "explainable") AND ("human-robot" OR "HRI" OR "social robot" OR "robot*") AND ("providing explanation" OR "explanation generation" OR "generate explanations" OR "answering questions")) OR (("Abstract": "XAR" OR "explainable autonomous robot*"OR "explainability" OR "interpretability" OR "explainable") AND ("human-robot" OR "HRI" OR "social robot" OR "robot*") AND ("providing explanation" OR "explanation generation" OR "generate explanations" OR "answering questions")) OR (("Author Keywords": "XAR" OR "explainable autonomous robot*"OR "explainability" OR "interpretability" OR "explainable") AND ("human-robot" OR "HRI" OR "social robot" OR "robot*") AND ("providing explanation" OR "explanation generation" OR "generate explanations" OR "answering questions"))))

\paragraph{Search String 4 (Scopus)} ( "XAR" OR "explainable autonomous robots" OR "explainable autonomous robot" OR "explainability" OR "explaining" OR "interpretability" OR "explainable" ) AND ( "human-robot" OR "HRI" OR "social robot" OR "robots" OR "robot" OR "robotic" OR "robotics" ) AND ( "providing explanation" OR "explanation generation" OR "generate explanations" OR "answering questions" )

\paragraph{Search String 5 (Springer Link)} ("explainability" OR "explainable" OR "interpretability") AND ("XAR" OR "explainable autonomous robots" OR "explainable autonomous robot (XAR)" OR "autonomous robot") AND ("human-robot" OR "HRI" OR "social robot" OR "robots" Or "robotics" OR "robot") AND ("algorithm" OR "framework" OR "system") AND ("providing explanation" OR "explanation generation" OR "generate explanations" OR "answering questions")

\section*{Acknowledgment}
This research has been partially funded by the Recovery, Transformation, and Resilience Plan, financed by the European Union (Next Generation) thanks to the TESCAC project (Traceability and Explainability in Autonomous Systems for improved Cybersecurity) granted by INCIBE to the University of León, and by grant PID2021-126592OB-C21 funded by MCIN/AEI/10.13039/501100011033
EDMAR (Explainable Decision Making in Autonomous Robots) project, PID2021-126592OB-C21 funded by MCIN/AEI/10.13039/501100011033 and by ERDF ''A way of making Europe''. David Sobrín-Hidalgo thanks the University of León for funding his doctoral studies.
% The preferred spelling of the word ``acknowledgment'' in American English is
% without an ``e'' after the ``g.'' Use the singular heading even if you have
% many acknowledgments. Avoid expressions such as ``One of us (S.B.A.) would
% like to thank $\ldots$ .'' Instead, write ``F. A. Author thanks $\ldots$ .'' In most
% cases, sponsor and financial support acknowledgments are placed in the
% unnumbered footnote on the first page, not here.

\bibliographystyle{IEEEtran}
\bibliography{cites.bib}

\begin{thebibliography}{10}
\providecommand{\url}[1]{#1}
\csname url@samestyle\endcsname
\providecommand{\newblock}{\relax}
\providecommand{\bibinfo}[2]{#2}
\providecommand{\BIBentrySTDinterwordspacing}{\spaceskip=0pt\relax}
\providecommand{\BIBentryALTinterwordstretchfactor}{4}
\providecommand{\BIBentryALTinterwordspacing}{\spaceskip=\fontdimen2\font plus
\BIBentryALTinterwordstretchfactor\fontdimen3\font minus \fontdimen4\font\relax}
\providecommand{\BIBforeignlanguage}[2]{{%
\expandafter\ifx\csname l@#1\endcsname\relax
\typeout{** WARNING: IEEEtran.bst: No hyphenation pattern has been}%
\typeout{** loaded for the language `#1'. Using the pattern for}%
\typeout{** the default language instead.}%
\else
\language=\csname l@#1\endcsname
\fi
#2}}
\providecommand{\BIBdecl}{\relax}
\BIBdecl

\bibitem{xarSurvey}
\BIBentryALTinterwordspacing
T.~Sakai and T.~Nagai, ``Explainable autonomous robots: a survey and perspective,'' \emph{Advanced Robotics}, vol.~36, no. 5-6, pp. 219--238, 2022. [Online]. Available: \url{https://doi.org/10.1080/01691864.2022.2029720}
\BIBentrySTDinterwordspacing

\bibitem{kumar2023study}
S.~Kumar, S.~Sarraf, A.~K. Kar, and P.~V. Ilavarasan, ``A study of explainable artificial intelligence: A systematic literature review of the applications,'' \emph{IoT, Big Data and AI for Improving Quality of Everyday Life: Present and Future Challenges: IOT, Data Science and Artificial Intelligence Technologies}, pp. 243--259, 2023.

\bibitem{vilone2020explainable}
G.~Vilone and L.~Longo, ``Explainable artificial intelligence: a systematic review,'' \emph{arXiv preprint arXiv:2006.00093}, 2020.

\bibitem{wallkotter2021explainable}
S.~Wallk{\"o}tter, S.~Tulli, G.~Castellano, A.~Paiva, and M.~Chetouani, ``Explainable embodied agents through social cues: a review,'' \emph{ACM Transactions on Human-Robot Interaction (THRI)}, vol.~10, no.~3, pp. 1--24, 2021.

\bibitem{anjomshoae2019explainable}
S.~Anjomshoae, A.~Najjar, D.~Calvaresi, and K.~Fr{\"a}mling, ``Explainable agents and robots: Results from a systematic literature review,'' in \emph{18th International Conference on Autonomous Agents and Multiagent Systems (AAMAS 2019), Montreal, Canada, May 13--17, 2019}.\hskip 1em plus 0.5em minus 0.4em\relax International Foundation for Autonomous Agents and Multiagent Systems, 2019, pp. 1078--1088.

\bibitem{moher_prisma_2010}
\BIBentryALTinterwordspacing
D.~Moher, A.~Liberati, J.~Tetzlaff, and D.~G. Altman, ``\BIBforeignlanguage{en}{Preferred reporting items for systematic reviews and meta-analyses: {The} {PRISMA} statement},'' \emph{\BIBforeignlanguage{en}{International Journal of Surgery}}, vol.~8, no.~5, pp. 336--341, 2010. [Online]. Available: \url{https://linkinghub.elsevier.com/retrieve/pii/S1743919110000403}
\BIBentrySTDinterwordspacing

\bibitem{method_kitchenham_2015}
\BIBentryALTinterwordspacing
B.~A. Kitchenham, D.~Budgen, and P.~Brereton, \emph{\BIBforeignlanguage{en}{Evidence-{Based} {Software} {Engineering} and {Systematic} {Reviews}}}, 0th~ed.\hskip 1em plus 0.5em minus 0.4em\relax Chapman and Hall/CRC, Nov. 2015. [Online]. Available: \url{https://www.taylorfrancis.com/books/9781482228663}
\BIBentrySTDinterwordspacing

\bibitem{Gonz_lez_Santamarta_2023}
\BIBentryALTinterwordspacing
M.~A. González-Santamarta, L.~Fernández-Becerra, D.~Sobrín-Hidalgo, A.~M. Guerrero-Higueras, I.~González, and F.~J.~R. Lera, \emph{Using Large Language Models for Interpreting Autonomous Robots Behaviors}.\hskip 1em plus 0.5em minus 0.4em\relax Springer Nature Switzerland, 2023, p. 533–544. [Online]. Available: \url{http://dx.doi.org/10.1007/978-3-031-40725-3\_45}
\BIBentrySTDinterwordspacing

\bibitem{han_building_2021}
\BIBentryALTinterwordspacing
Z.~Han, D.~Giger, J.~Allspaw, M.~S. Lee, H.~Admoni, and H.~A. Yanco, ``Building the {Foundation} of {Robot} {Explanation} {Generation} {Using} {Behavior} {Trees},'' \emph{ACM Transactions on Human-Robot Interaction}, vol.~10, no.~3, pp. 26:1--26:31, Jul. 2021. [Online]. Available: \url{https://dl.acm.org/doi/10.1145/3457185}
\BIBentrySTDinterwordspacing

\bibitem{sakai_explainable_2022}
\BIBentryALTinterwordspacing
T.~Sakai and T.~Nagai, ``\BIBforeignlanguage{en}{Explainable autonomous robots: a survey and perspective},'' \emph{\BIBforeignlanguage{en}{Advanced Robotics}}, vol.~36, no. 5-6, pp. 219--238, Mar. 2022. [Online]. Available: \url{https://www.tandfonline.com/doi/full/10.1080/01691864.2022.2029720}
\BIBentrySTDinterwordspacing

\bibitem{halilovic2023visuo}
A.~Halilovic and F.~Lindner, ``Visuo-textual explanations of a robot's navigational choices,'' in \emph{Companion of the 2023 ACM/IEEE International Conference on Human-Robot Interaction}, 2023, pp. 531--535.

\bibitem{explainingInTime}
\BIBentryALTinterwordspacing
T.~Arnold, D.~Kasenberg, and M.~Scheutz, ``Explaining in time: Meeting interactive standards of explanation for robotic systems,'' \emph{J. Hum.-Robot Interact.}, vol.~10, no.~3, jul 2021. [Online]. Available: \url{https://doi.org/10.1145/3457183}
\BIBentrySTDinterwordspacing

\bibitem{fernandez2023accountability}
L.~Fern{\'a}ndez-Becerra, M.~A. Gonz{\'a}lez-Santamarta, D.~Sobr{\'\i}n-Hidalgo, {\'A}.~M. Guerrero-Higueras, F.~J.~R. Lera, and V.~M. Olivera, ``Accountability and explainability in robotics: A proof of concept for ros 2-and nav2-based mobile robots,'' in \emph{Computational Intelligence in Security for Information Systems Conference}.\hskip 1em plus 0.5em minus 0.4em\relax Springer, 2023, pp. 3--13.

\bibitem{sakai_implementation_2023}
\BIBentryALTinterwordspacing
T.~Sakai, T.~Nagai, and K.~Abe, ``Implementation and {Evaluation} of {Algorithms} for {Realizing} {Explainable} {Autonomous} {Robots},'' \emph{IEEE Access}, pp. 1--1, 2023. [Online]. Available: \url{https://ieeexplore.ieee.org/document/10210548/}
\BIBentrySTDinterwordspacing

\bibitem{borgo_towards_2018}
\BIBentryALTinterwordspacing
R.~Borgo, M.~Cashmore, and D.~Magazzeni, ``Towards {Providing} {Explanations} for {AI} {Planner} {Decisions},'' 2018. [Online]. Available: \url{https://arxiv.org/abs/1810.06338}
\BIBentrySTDinterwordspacing

\bibitem{schardt_utilization_2007}
\BIBentryALTinterwordspacing
C.~Schardt, M.~B. Adams, T.~Owens, S.~Keitz, and P.~Fontelo, ``\BIBforeignlanguage{en}{Utilization of the {PICO} framework to improve searching {PubMed} for clinical questions},'' \emph{\BIBforeignlanguage{en}{BMC Medical Informatics and Decision Making}}, vol.~7, no.~1, p.~16, Dec. 2007. [Online]. Available: \url{https://bmcmedinformdecismak.biomedcentral.com/articles/10.1186/1472-6947-7-16}
\BIBentrySTDinterwordspacing

\bibitem{zhang_validation_2011}
\BIBentryALTinterwordspacing
H.~Zhang, M.~A. Babar, and P.~Tell, ``\BIBforeignlanguage{en}{Identifying relevant studies in software engineering},'' \emph{\BIBforeignlanguage{en}{Information and Software Technology}}, vol.~53, no.~6, pp. 625--637, Jun. 2011. [Online]. Available: \url{https://linkinghub.elsevier.com/retrieve/pii/S0950584910002260}
\BIBentrySTDinterwordspacing

\bibitem{DybraQualityEvidences}
\BIBentryALTinterwordspacing
T.~Dyb\r{a} and T.~Dings\o{}yr, ``Strength of evidence in systematic reviews in software engineering,'' in \emph{Proceedings of the Second ACM-IEEE International Symposium on Empirical Software Engineering and Measurement}, ser. ESEM '08.\hskip 1em plus 0.5em minus 0.4em\relax New York, NY, USA: Association for Computing Machinery, 2008, p. 178–187. [Online]. Available: \url{https://doi.org/10.1145/1414004.1414034}
\BIBentrySTDinterwordspacing

\bibitem{halilovic2023influence}
A.~Halilovic and S.~Krivic, ``The influence of a robot’s personality on real-time explanations of its navigation,'' in \emph{International Conference on Social Robotics}.\hskip 1em plus 0.5em minus 0.4em\relax Springer, 2023, pp. 133--147.

\bibitem{han2023communicating}
Z.~Han and H.~Yanco, ``Communicating missing causal information to explain a robot’s past behavior,'' \emph{ACM Transactions on Human-Robot Interaction}, vol.~12, no.~1, pp. 1--45, 2023.

\bibitem{rosenthal2022impact}
S.~Rosenthal, P.~Vichivanives, and E.~Carter, ``The impact of route descriptions on human expectations for robot navigation,'' \emph{ACM Transactions on Human-Robot Interaction (THRI)}, vol.~11, no.~4, pp. 1--19, 2022.

\bibitem{cruz2023explainable}
F.~Cruz, R.~Dazeley, P.~Vamplew, and I.~Moreira, ``Explainable robotic systems: Understanding goal-driven actions in a reinforcement learning scenario,'' \emph{Neural Computing and Applications}, vol.~35, no.~25, pp. 18\,113--18\,130, 2023.

\bibitem{hu2023explainable}
S.~Hu and T.~Nagai, ``Explainable autonomous robots in continuous state space based on graph-structured world model,'' \emph{Advanced Robotics}, vol.~37, no.~16, pp. 1025--1041, 2023.

\bibitem{BOGATARKAN_ERDEM_2020}
A.~BOGATARKAN and E.~ERDEM, ``Explanation generation for multi-modal multi-agent path finding with optimal resource utilization using answer set programming,'' \emph{Theory and Practice of Logic Programming}, vol.~20, no.~6, p. 974–989, 2020.

\bibitem{mualla2022quest}
Y.~Mualla, I.~Tchappi, T.~Kampik, A.~Najjar, D.~Calvaresi, A.~Abbas-Turki, S.~Galland, and C.~Nicolle, ``The quest of parsimonious xai: A human-agent architecture for explanation formulation,'' \emph{Artificial intelligence}, vol. 302, p. 103573, 2022.

\bibitem{zakershahrak2020online}
M.~Zakershahrak, Z.~Gong, N.~Sadassivam, and Y.~Zhang, ``Online explanation generation for planning tasks in human-robot teaming,'' in \emph{2020 IEEE/RSJ International Conference on Intelligent Robots and Systems (IROS)}.\hskip 1em plus 0.5em minus 0.4em\relax IEEE, 2020, pp. 6304--6310.

\bibitem{mota2021answer}
T.~Mota and M.~Sridharan, ``Answer me this: constructing disambiguation queries for explanation generation in robotics,'' in \emph{2021 IEEE International Conference on Development and Learning (ICDL)}.\hskip 1em plus 0.5em minus 0.4em\relax IEEE, 2021, pp. 1--8.

\bibitem{gong2018behavior}
Z.~Gong and Y.~Zhang, ``Behavior explanation as intention signaling in human-robot teaming,'' in \emph{2018 27th IEEE International Symposium on Robot and Human Interactive Communication (RO-MAN)}.\hskip 1em plus 0.5em minus 0.4em\relax IEEE, 2018, pp. 1005--1011.

\bibitem{schroeter2022introspectionbasedexplainablereinforcementlearning}
\BIBentryALTinterwordspacing
N.~Schroeter, F.~Cruz, and S.~Wermter, ``Introspection-based explainable reinforcement learning in episodic and non-episodic scenarios,'' 2022. [Online]. Available: \url{https://arxiv.org/abs/2211.12930}
\BIBentrySTDinterwordspacing

\bibitem{gao2020joint}
X.~Gao, R.~Gong, Y.~Zhao, S.~Wang, T.~Shu, and S.-C. Zhu, ``Joint mind modeling for explanation generation in complex human-robot collaborative tasks,'' in \emph{2020 29th IEEE international conference on robot and human interactive communication (RO-MAN)}.\hskip 1em plus 0.5em minus 0.4em\relax IEEE, 2020, pp. 1119--1126.

\bibitem{brandao2021towards}
M.~Brandao, G.~Canal, S.~Krivi{\'c}, and D.~Magazzeni, ``Towards providing explanations for robot motion planning,'' in \emph{2021 IEEE International Conference on Robotics and Automation (ICRA)}.\hskip 1em plus 0.5em minus 0.4em\relax IEEE, 2021, pp. 3927--3933.

\bibitem{chen2020towards}
S.~Chen, K.~Boggess, and L.~Feng, ``Towards transparent robotic planning via contrastive explanations,'' in \emph{2020 IEEE/RSJ International Conference on Intelligent Robots and Systems (IROS)}.\hskip 1em plus 0.5em minus 0.4em\relax IEEE, 2020, pp. 6593--6598.

\bibitem{olivares2021knowledge}
A.~Olivares-Alarcos, S.~Foix, and G.~Aleny{\`a}, ``Knowledge representation for explainability in collaborative robotics and adaptation,'' 2021.

\bibitem{brandao2021explaining}
M.~Brandao, A.~Coles, and D.~Magazzeni, ``Explaining path plan optimality: Fast explanation methods for navigation meshes using full and incremental inverse optimization,'' in \emph{Proceedings of the International Conference on Automated Planning and Scheduling}, vol.~31, 2021, pp. 56--64.

\bibitem{rosenthal2016verbalization}
S.~Rosenthal, S.~P. Selvaraj, and M.~M. Veloso, ``Verbalization: Narration of autonomous robot experience.'' in \emph{IJCAI}, vol.~16, 2016, pp. 862--868.

\end{thebibliography}

\begin{IEEEbiography}[{\includegraphics[width=1in,height=1.25in,clip,keepaspectratio]{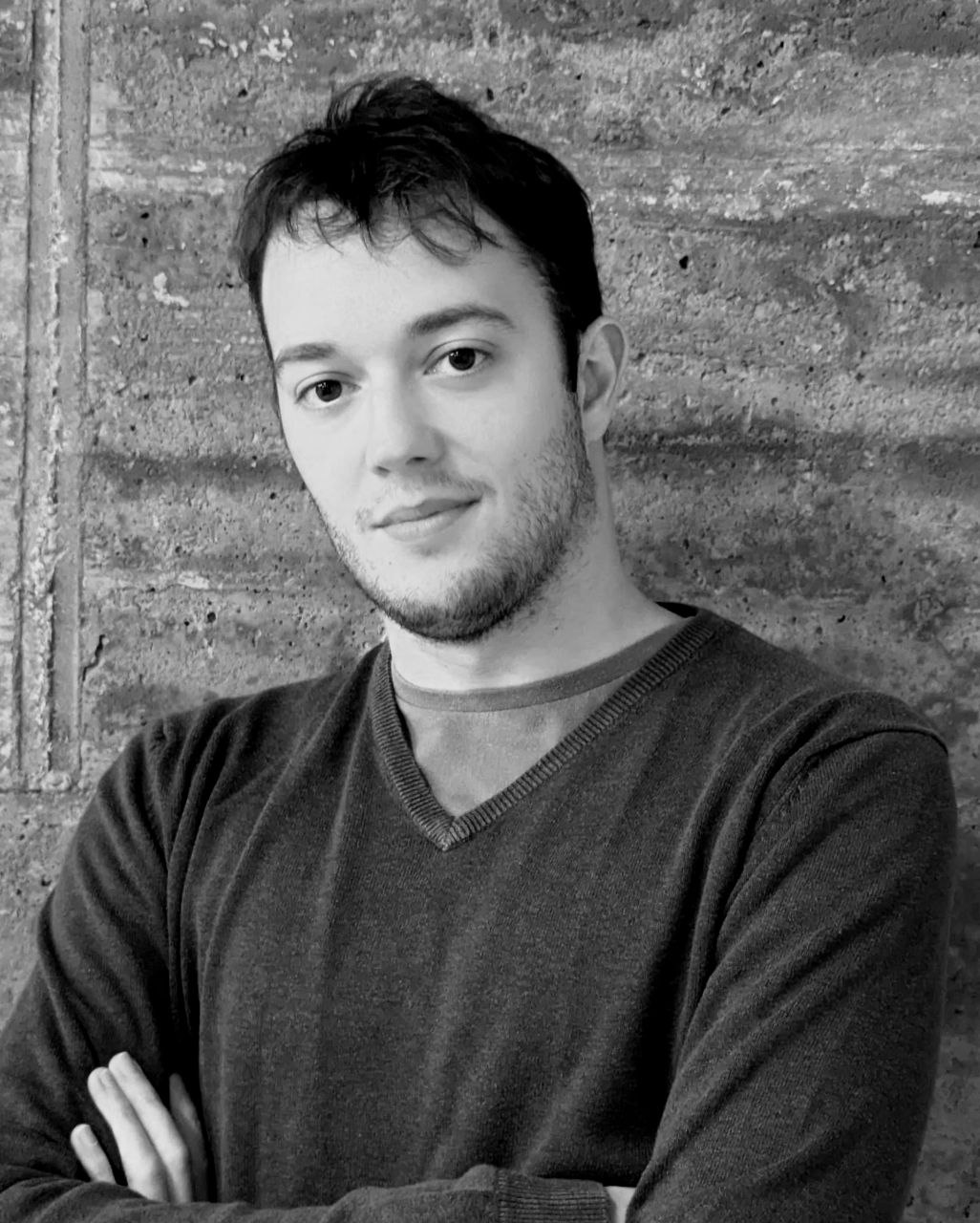}}]{David Sobrín-Hidalgo} received the B.S. degree in computer engineering from the University of León (Spain) in 2018, and the M.S. degree in cybersecurity research from the same institution in 2020. During his master's studies, he completed an external internship in 2020 with the Robotics Group of the University of Leon. His research interests include working with robotic systems and Large Language Models (LLMs).

In 2022, he began his Ph.D. studies in the Production and Computing Engineering program at the University of León, where he is a member of the Robotics Group. His doctoral research focuses on explainability and cybersecurity in robotics. Since beginning his Ph.D. studies, he has participated in several research projects and contributed to multiple scientific publications in his field.
\end{IEEEbiography}

\begin{IEEEbiography}[{\includegraphics[width=1in,height=1.25in,clip,keepaspectratio]{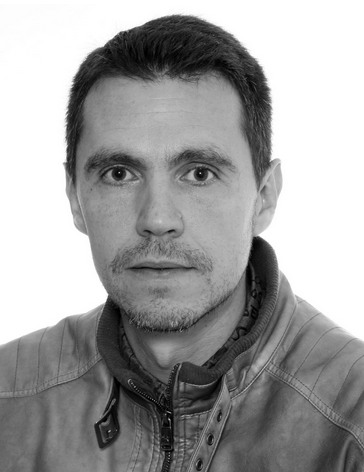}}]{Ángel M. Guerrero-Higueras} 
Ángel Manuel Guerrero-Higueras, an IT engineer, has over two decades of experience in both the private sector (at leading companies such as Telefónica and Indra) and academia. He holds a PhD from the University of León, where he is currently a Professor, specializing in cybersecurity and robotics. As a researcher in the Atmospheric Physics Group (from 2011 to 2013) and in the Institute of Applied Sciences for Cybersecurity (from 2016 to 2018), he has supervised 5 PhD theses and published more than 80 works. He is currently an active member of the Robotics Group at the University of León.
\end{IEEEbiography}

%If you do not have or do not want to include a photo, you can use IEEEbiographynophoto as shown below:

\begin{IEEEbiography}[{\includegraphics[width=1in,height=1.25in,clip,keepaspectratio]{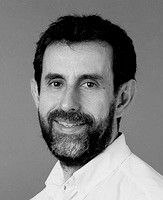}}]{Vicente Matellán-Olivera} Dr. Vicente Matellán Olivera got his PhD in Computer Science from the Technical University of Madrid in 1993. He was Assistant Professor at Universidad Carlos III de Madrid (Spain) (1993-1999). Associate Professor at Universidad Rey Juan Carlos (Spain) from 1999-2008. In 2008 he joined Universidad de León (León), where he still serves as Full Professor in the Mechanical, Computer and Aerospace Engineering Department. His main research interests have to do with robotics, artificial intelligence, and cybersecurity areas where he has made more than 250 contributions in journals, books, and conferences.
\end{IEEEbiography}
\EOD
\end{document}